\documentclass{l4dc2025}

\title[Learning STL with Statistical Guarantees]{Learning Temporal Logic Predicates from Data with Statistical Guarantees}
\usepackage{times}

\author{%
	\Name{Emi Soroka} \Email{esoroka@stanford.edu}\\
	\addr Department of Aeronautics and Astronautics
		Stanford University
	\AND
	\Name{Rohan Sinha} \Email{rhnsinha@stanford.edu}\\
	\addr Department of Aeronautics and Astronautics
		Stanford University
	\AND
	\Name{Sanjay Lall} \Email{lall@stanford.edu}\\
	\addr Department of Electrical Engineering
		Stanford University
}

\usepackage{booktabs}       
\usepackage{nicefrac}       
\usepackage{microtype}      
\usepackage{xcolor}         
\usepackage{wrapfig}
\usepackage{verbatimbox}

\newcommand{\R}{\mathbb{R}}
\newcommand{\Dd}{\mathcal{D}}
\newcommand{\Gg}{\mathcal{G}}
\newcommand{\Tt}{\mathcal{T}}
\newcommand{\train}{\text{train}}
\newcommand{\calib}{\text{cal}}
\newcommand{\test}{\text{test}}
\newcommand{\val}{\text{val}}
\newcommand{\obs}{\mathop{obs}}
\newcommand{\Prob}{\mathbf{P}}

\newcommand{\Eventually}{\lozenge}

\begin{document}

\maketitle

\begin{abstract}

Temporal logic rules are often used in control and robotics to provide structured, human-interpretable descriptions of trajectory data. These rules have numerous applications including safety validation using formal methods, constraining motion planning among autonomous agents, and classifying data. However, existing methods for learning temporal logic predicates from data do not provide assurances about the correctness of the resulting predicate.
We present a novel method to learn temporal logic predicates from data with finite-sample correctness guarantees. Our approach leverages expression optimization and conformal prediction to learn predicates that correctly describe future trajectories under mild statistical assumptions. We provide experimental results showing the performance of our approach on a simulated trajectory dataset and perform ablation studies to understand how each component of our algorithm contributes to its performance.

\end{abstract}


\section{Introduction}
This paper presents a novel algorithm that takes a dataset of trajectories and learns a signal temporal logic predicate which describes future (unseen) trajectories, under mild assumptions, with high probability.
Many systems of interest, such as vehicle traffic or interactions between humans and autonomous robots, exhibit complex, time-dependent behavior that can be modeled by signal temporal logic (STL): a mathematical language for expressing logical predicates over time series data. However, STL predicates are difficult for humans to specify by hand (\cite{hahn2022formal}) and may change as a system evolves over time. Thus in many applications, it is desirable to learn temporal logic predicates from data. Mining such predicates can yield new insights into system behavior (\cite{pigozzi2021mining}) and failure modes (\cite{corso2020interpretable}), as well as providing empirically-derived logical constraints for use in motion planning algorithms (\cite{reactive_synthesis}).

Because temporal logic lends itself to formal verification of system properties (e.g. safety or liveness proofs) (\cite{nenzi2023learning}), it is critically important to develop predicates that accurately model the system under verification. Although many algorithms have been developed to learn temporal predicates from data, safety-critical predicates must still be written by human experts (\cite{maierhofer2022formalization, 8972130}) because currently available algorithms do not provide any guarantees on the correctness of learned predicates.
We present a novel approach for learning temporal logic predicates with finite-sample correctness guarantees using conformal prediction. Under mild assumptions, our algorithm learns predicates with a guaranteed probability of correctness on unseen test data. Because we apply expression optimization to learn an optimal predicate from a set of smaller STL features, we additionally present a novel penalty function over expression trees that prevents trivial expressions (those that simplify to Boolean \verb|true| or \verb|false|) from being generated. Finally, we provide empirical results showing the effectiveness of our approach.
\section{Related Work}
The task of learning the parameters or structure of a temporal logic predicate has been extensively studied in the literature. 
(\cite{leung2023backpropagation}) use backpropagation to learn numeric parameters of STL predicates via gradient descent. Supervised learning for prediction tasks (\cite{qin2020clairvoyant}), unsupervised classification (\cite{vazquez2017logical}), decision trees (\cite{bombara2021decision}), randomized optimization (\cite{pigozzi2021mining}) and many other paradigms have been applied to this problem.
We apply expression optimization, a class of randomized algorithms for learning tree structures, to fit STL predicates to data. This approach provides the advantages of being simple to implement using published software and highly efficient in practice.


(\cite{nenzi2018robust}) presents a two-step approach consisting of a novel genetic algorithm for learning the structure of STL predicates and a parameter fitting step over simulated robustness values. This tactic is similar to the trajectory prediction step in our algorithm (Figure 2), which generates predicted robustness values for each trajectory in the dataset. One major difference is that Nenzi et al. learns a predicate that distinguishes normal from anomalous trajectories using labeled data. Thus the metrics they present are the mean misclassification rate as well as the false positive and false negative rates.

(\cite{roy2023learning}) studies the one-class classification problem: learning a predicate that describes the entire dataset. However, while our paper is concerned with the statistical validity of learned STL predicates, Roy et al. focus on learning one-class LTL predicates efficiently, investigating inference time as a function of a complexity parameter in their algorithm.

Conformal methods have also seen increasing popularity, finding applications in robotic autonomy, where safety violations can have serious real-world consequences, and in controlling error rates for safe machine learning (\cite{angelopoulos2023conformal}). Conformalized quantile regression (CQR), which we apply in this paper, was developed in 2019 by 
(\cite{romano2019conformalized}) as a method of producing tighter conformal confidence intervals for real-valued predictors.
CQR has since been applied in in many problem domains including a predictive monitoring application relevant to our work: given a system and a predicate $\phi$ that describes desirable or safe system trajectories, what is the probability that future (unseen) trajectories satisfy $\phi$? This approach was first proposed in 2020 for temporal logic monitoring of ARIMA processes (\cite{qin2020clairvoyant}). Building on this work, 
(\cite{qin2022statistical}) use a surrogate model of the system being monitored to make predictions. Finally, two simultaneous publications extended this monitoring approach to Markovian systems and black-box prediction algorithms, respectively (\cite{lindemann2023conformal, cairoli2023conformal}). Our work leverages theoretical results from these two publications.
\section{Background}
Because this paper combines tools from statistics and formal verification, we provide a brief overview of both.

\textbf{Signal temporal logic (STL)} is a language for expressing logical statements over real-valued time series data \cite{koymans1990specifying, maler2004monitoring}. STL uses Boolean predicates (\textit{And}, \textit{Or}, \textit{Implies}, \textit{Not}) and temporal predicates (\textit{Eventually} ($\Eventually$), \textit{Always} ($\square$), \textit{Until} ($\mathcal{U}$)) to construct arbitrarily complex descriptions of trajectory data. These descriptions can then be evaluated over trajectories, yielding the Boolean values \verb|true| or \verb|false|. We consider STL predicates over finite trajectories of length $T$. We use $\Tt$ to describe the space of trajectories $x = (x_1,\dots,x_T)$, where $x_t$ is the system state at time $t$. If predicate $\phi$ is true for trajectory $x$, we say $x\models\phi$.

%
%
%
STL also admits a notion of real-valued robustness (\cite{donze2010robust}). The robustness can be defined recursively for arbitrarily complex STL expressions; we use the standard definitions given in (\cite{qin2020clairvoyant}). In this paper, we define  $\rho_\phi: \Tt\rightarrow\R$ as the robustness function for an STL predicate $\phi$. 
Some useful examples are:
\begin{align}
\begin{split}
\rho_{\Eventually_{[1,T]}(g(x_t) \geq 0)} &= \max_{t\in [1,T]} g(x_t)
\\
\rho_{\neg\phi}(x) &= -\rho_\phi(x)
\label{eq:robustness-example}
\\
\rho_{\phi_1\wedge\phi_2}(x) &= \min\{\rho_{\phi_1}(x), \rho_{\phi_2}(x)\}
\\
\rho_{\phi_1\vee\phi_2}(x) &= \max\{\rho_{\phi_1}(x), \rho_{\phi_2}(x)\}.
\end{split}
\end{align}
The robustness value measures how well a trajectory $x$ satisfies a given predicate $\phi$, defined such that $x\models\phi$ if $\rho_{\phi}(x) > 0$ \cite{lindemann2023conformal}.
We only consider temporal operators over the full length of the trajectory, so we drop the $[1,T]$ subscript for readability.



\textbf{Conformal prediction} is a method of computing confidence sets for black-box predictors with finite-sample coverage guarantees (\cite{vovk2005algorithmic}). Intuitively, conformal methods work by making the assumption that the test dataset is drawn from the same distribution as the training and calibration data. Thus, after training a predictor we only need to evaluate our prediction error on the calibration dataset and use this information to compute confidence sets for predictions over the test data. Consider a dataset $\Dd$ containing $N$ pairs of observations and labels $(x^{(i)}, y^{(i)})$. Given a new observation $x^{(N+1)}$, we wish to find a confidence set $C(x^{(N+1)})$ that contains the (unknown) label $y^{(N+1)}$ with high probability:
\begin{equation}
\Prob\left(y\in C(x)\right)\geq 1-\alpha
\label{eq:coverage}
\end{equation}
(where $\alpha$ is a user-specified parameter). Conformal prediction uses a \textit{nonconformity score}: a function that measures how well each data point ``conforms'' to the rest of the dataset, to compute $C(x)$. The choice of nonconformity score impacts the efficiency of the resulting confidence sets but not their validity. In conformalized regression, the absolute difference $z^{(i)} = |y^{(i)}-\hat y^{(i)}|$ is used (\cite{lei2018distribution}); however other score functions are used for other prediction methods (\cite{knnconformal2011, romano2019conformalized}). 

We apply \textit{split conformal quantile regression} (CQR) (\cite{romano2019conformalized}), which uses a calibration dataset~$\Dd_{\calib}$ to conformalize a predictor $f_\alpha:\mathcal{X}\rightarrow\R$ that estimates $q_{\alpha}$, the $\alpha$-th quantile of $y$. The resulting predicted confidence intervals have the desired coverage property \eqref{eq:coverage} when the data points are i.i.d. (\cite{romano2019conformalized}).\footnote{The i.i.d. assumption can be relaxed to exchangeability: i.e. the joint distribution of $(x_1,\dots,x_n)$ is invariant under permutation of the indices $\{1,\dots,n\}$.} For convenience, we define $f^1$ and $f^2$ such that for a trajectory $x\in\Tt$ and its robustness value $y\in\R$ evaluated over some STL predicate:
\begin{align*}
f^1(x) &= \hat q_{\nicefrac{\alpha}{2}} \quad \nicefrac{\alpha}{2}\text{-th conformalized quantile predictor of }y,
\\
f^2(x) &= \hat q_{1-\nicefrac{\alpha}{2}} \quad (1-\nicefrac{\alpha}{2})\text{-th conformalized quantile predictor of }y.
\end{align*}

%
Prior work has applied CQR to predict confidence intervals (CIs) for the robustness $\rho_\phi$ of a given STL predicate $\phi$. 
(\cite{lindemann2023conformal}) train a predictor to output a CI $[l, h]$ such that $ \Prob\left(\rho_\phi\left(x\right) \in [l,h]\right)\geq 1-\alpha $ on a test point $x$. If the CI $[l,h]$ contains only positive robustness values, then $\phi$ correctly describes the data with probability at least $1- \alpha$. (\cite{zhao2024robustconformalpredictionstl}) extends this approach to account for distribution shift in real-world applications. We extend the ideas of conformal prediction for STL in a different direction: solving the inverse problem of identifying a predicate $\phi$ that is correct with high probability. To do so, we must efficiently compose predicates in our expression optimization step. 
(\cite{cairoli2023conformal}) develop \textit{calibrated interval arithmetic} to perform this task, which we describe here. Given predictors $\phi_m$ and $\phi_n$ and robustness CIs $[f^1_{\phi_m} (x),f^2_{\phi_m} (x)]$, $[f^1_{\phi_n} (x), f^2_{\phi_n} (x)]$, we can approximate the combined CI of $\phi_m\wedge\phi_n$ as
\begin{equation}
[l_{\phi_m\wedge\phi_n}, h_{\phi_m\wedge\phi_n}] \subseteq [\min\{l_{\phi_m},l_{\phi_n}\} \min\{h_{\phi_m},h_{\phi_n}\}],
\label{eq:combined-and}
\end{equation}
and the combined confidence interval of $\phi_m\vee\phi_n$ as
\begin{equation}
[l_{\phi_m\vee\phi_n}, h_{\phi_m\vee\phi_n}] \subseteq [\max\{l_{\phi_m},l_{\phi_n}\}, \max\{h_{\phi_m},h_{\phi_n}\}].
\label{eq:combined-or}
\end{equation}
Since $\rho_{\neg\phi}(x) = -\rho_{\phi}(x)$, the CI of $\neg\phi$ is $[\min\{-l_{\phi}, -h_{\phi}\}, \max\{-l_{\phi}, -h_{\phi}\}]$.
This approximation method requires computing a new conformal adjustment $Q_{1-\alpha}$ for the CQR predictor corresponding to the combined predicate.

\section{Problem Statement}
We are interested in \textit{learning} a predicate $\phi$ from a dataset of trajectories $\Dd := \{x^{(1)},\dots,x^{(N)}\}$ such that the corresponding robustness confidence interval $[l_\phi,h_\phi]$ is \textit{tight}, \textit{positive} ($0\leq l_\phi<h_\phi$), and contains the true robustness value of future trajectories with high probability.

Our algorithm leverages calibrated interval arithmetic (\cite{cairoli2023conformal}) and expression optimization to efficiently learn an STL predicate $\phi^\star$ with a desired robustness confidence interval $[l,h]$. We then compute a CQR predictor for the robustness of $\phi^\star$. Thus, our algorithm provides both an STL predicate describing the dataset and a statistically valid confidence interval predictor for the robustness of future trajectories. The novel contribution of our work is not an optimization algorithm or predictor; rather it is a framework for integrating these components using conformal prediction to learn predicates with statistical guarantees.

\begin{figure}[t]
	\centering
	\includegraphics[width=0.9\linewidth]{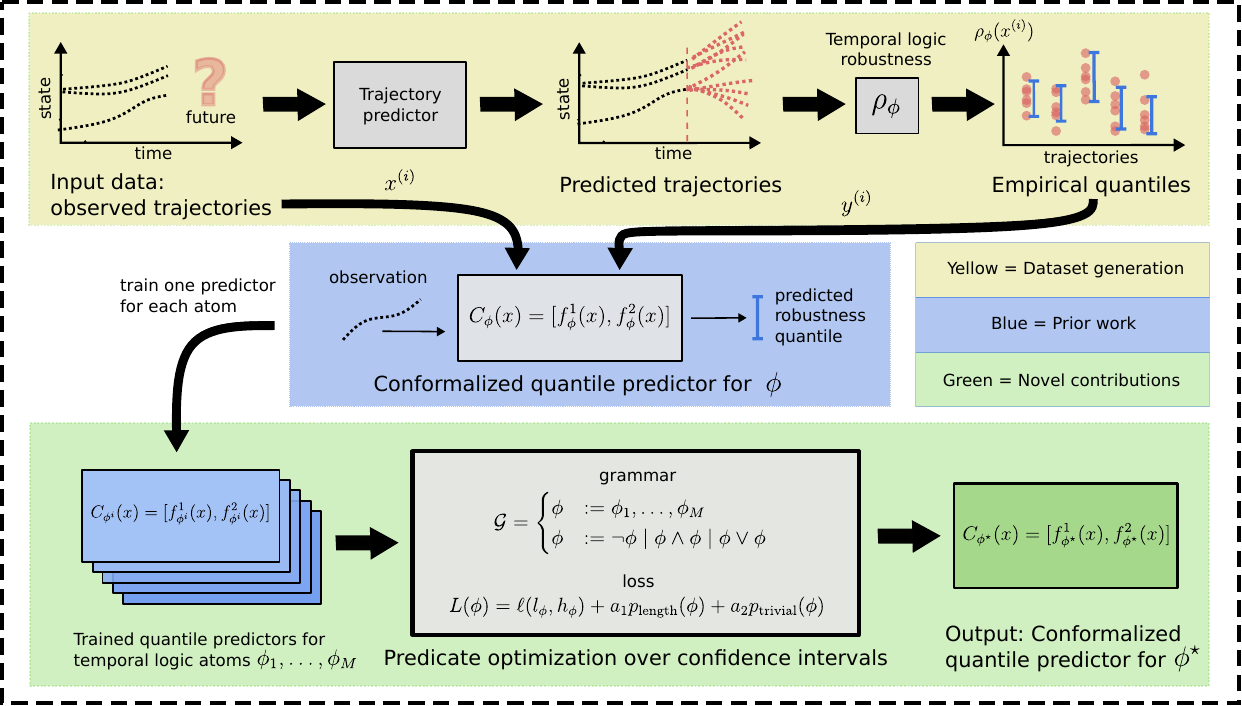}
	\caption{Block diagram showing the sequence of components in our algorithm. Steps shaded in yellow are part of the dataset generation process. Blue represent prior work and green represents novel contributions.}
	\label{fig:system}
 \vspace{-1em}
\end{figure}

\paragraph*{Dataset} We split $\Dd$ into $\Dd_{\train}$, $\Dd_{\calib 1}$, $\Dd_{\test}$, $\Dd_{\calib 2}$, and $\Dd_{\val}$. The five-way split is necessary because, as shown in Figure \ref{fig:system}, we first train conformalized quantile predictors for each temporal logic atom (using $\Dd_{\train}$ and $\Dd_{\calib 1}$). Then, treating $\Dd_{\test}$ as the test dataset for each atom's predictor, we evaluate the atoms, run the optimization step using the results of this evaluation, and finally conformalize the resulting predicate $\phi^\star$ using the independent calibration set $\Dd_{\calib 2}$ before evaluating its performance on the unseen dataset $\Dd_{\val}$. Because $\phi^\star$ is learned using the quantile predictors for $\phi_1,\dots,\phi_m$, which are conformalized using $\Dd_{\calib 1}$, $\phi^\star$ is a function of this first calibration set. Thus a separate calibration set is required to determine confidence intervals for $\phi^\star$ to maintain the i.i.d. assumption for CQR.

Although $\Dd$ contains complete trajectories over $t=\{0,\dots,T\}$, we are interested in predicting the STL robustness of future trajectories. Thus the input to our trajectory predictor is an observation $\obs(x^{(i)})$ containing partial or noisy trajectory data. This simulates a robotics application in which we wish to predict the robustness of an entire trajectory given a partial observation. We denote the space of observations $\mathcal{O}$. The user may select any trajectory predictor where the predictions remain exchangeable between the training, test, and calibration datasets.

\paragraph*{Predicate} We consider $M$ \textit{atoms} $\phi_1,\dots,\phi_M$ which are simple, meaningful predicates that will be combined to form the desired predicate $\phi$. The choice of $\phi_1,\dots,\phi_M$ and their parameters is a feature engineering task informed by our knowledge of the underlying system; for example, a reasonable choice is to indicate whether an agent is inside an unsafe region of the state space. While selecting atoms restricts the format of the resulting predicate, it also allows users to control the format of learned STL predicates and answer specific questions about the dataset. We expect this to be useful in situations where the environment is well-known, such as a road network or robotics facility in an autonomous navigation context. Our method could also be extended by using the expression optimization step to modify atom parameters.]
We also note that the only prerequisite to apply CQR is exchangeability of the data. Thus while the choice of atoms affects how meaningful a predicate is for a particular application, a poor choice cannot cause a violation of the conformal guarantee.

\paragraph*{Predictor} A \textit{quantile predictor} at level $\alpha$ is a predictor that outputs the estimated $\alpha$-th quantile of the target value. We denote the $\nicefrac{\alpha}{2}$ and $1-\nicefrac{\alpha}{2}$ conformalized quantile predictors for robustness $\rho_\phi$ as $f^1_\phi, f^2_\phi: \mathcal{O}\rightarrow \R$.
Our algorithm trains and conformalizes predictors $f_\phi^1$ and $f_\phi^2$ for each predicate atom. The user can select an appropriate prediction algorithm for their application; conformal methods do not require any underlying assumptions on predictor structure. We use $f_\phi^1(x)$ and $f_\phi^2(x)$ to refer to conformalized predictors which take as input an observation of trajectory $x$.

\paragraph*{Expression optimization}
We cast our problem as an expression optimization problem over expressions constructed from the pre-selected atoms $\phi_1,\dots,\phi_M$ and Boolean functions
\begin{equation}
\Gg_m = \begin{cases}
\phi &:= \phi_1,\dots,\phi_M
\\
\phi &:= \neg\phi\mid \phi \wedge \phi\mid \phi \vee \phi
\end{cases}.
\label{eq:grammar}
\end{equation}
We define a loss function $\ell(h_\phi,l_\phi)$ over a confidence interval $[l_\phi, h_\phi]$ for $\rho_{\phi}$ and solve
\begin{equation}
\underset{\phi}{\text{minimize}}\quad \frac{1}{n}\sum_{i=1}^N\ell\left(f^1_\phi(x^{(i)}), f^2_\phi(x^{(i)})\right).
\label{eq:exprprob}
\end{equation}
By optimizing over the predicted robustness confidence intervals of $\phi_1,\dots,\phi_M$ instead of the robustness values $\rho_{\phi_i}(x^{(i)})$ (which represent a larger dataset to evaluate at each iteration and may contain outliers), we are able to efficiently find a more robust optimal predicate $\phi^\star$. This problem formulation is one of the novel contributions of our work.

Because the final step of our algorithm is to conformalize the CI predictors for $\rho_{\phi^\star}$, the following property \eqref{eq:algorithm-guarantee} holds for an unseen trajectory $x^{(N+1)}$
\begin{equation}
\Prob\left(\rho_{\phi^\star}(x^{(N+1)})\in [f^1_{\phi^\star}(x^{(N+1)}), f^2_{\phi^\star}(x^{(N+1)})]\right)\geq 1-\alpha.
\label{eq:algorithm-guarantee}
\end{equation}

In general, Problem \eqref{eq:exprprob} is a difficult combinatorial optimization problem. However, solutions can be found using expression optimization methods such as genetic programming (GP), an algorithm that evolves a candidate expression from a population of random trees using crossover operations (swapping subtrees between trees $i$ and $j$) and mutation (making a random modification to a tree) to explore the solution space (\cite{koza1994genetic}). GP is part of a larger class of randomized expression optimization techniques; our paper tests several methods from this class to show that our approach is agnostic to the choice of optimizer. While these methods do not provide any guarantees of finding a globally optimal expression, or even a nontrivial expression, they are highly effective in practice and have been applied in multiple other papers on fitting STL predicates (\cite{qin2022statistical, pigozzi2021mining}). As part of our contributions, we discuss loss functions over STL robustness and develop a novel penalty over STL expression trees that significantly reduces the likelihood of the optimizer returning a trivial predicate (one that simplifies to Boolean \texttt{true} or \texttt{false}). 

\subsection*{Combining STL predicates} Using \eqref{eq:combined-and} and \eqref{eq:combined-or} to compose confidence intervals, we can approximate a CI $[l_\phi,h_\phi]$ for any $\phi$ constructed using the simplified grammar $\Gg_m$. This allows us to perform the (potentially expensive) steps of conformalizing $f^1_{\phi_i}$ and $f^2_{\phi_i}$ for each atom, evaluate them over $\Dd_{\train}$ once, then rapidly iterate over possible combinations. Once a solution $\phi^\star$ is found, we conformalize $f^1_{\phi^\star}, f^2_{\phi^\star}$ using the second calibration set. This step is necessary because \eqref{eq:combined-and} and \eqref{eq:combined-or} are only approximations to the true interval, thus the conformal adjustment must be re-computed with an independent calibration set to ensure the guarantee \eqref{eq:algorithm-guarantee} holds.

\begin{wrapfigure}{R}{0.5\textwidth}
	\centering
	\includegraphics[width=\linewidth]{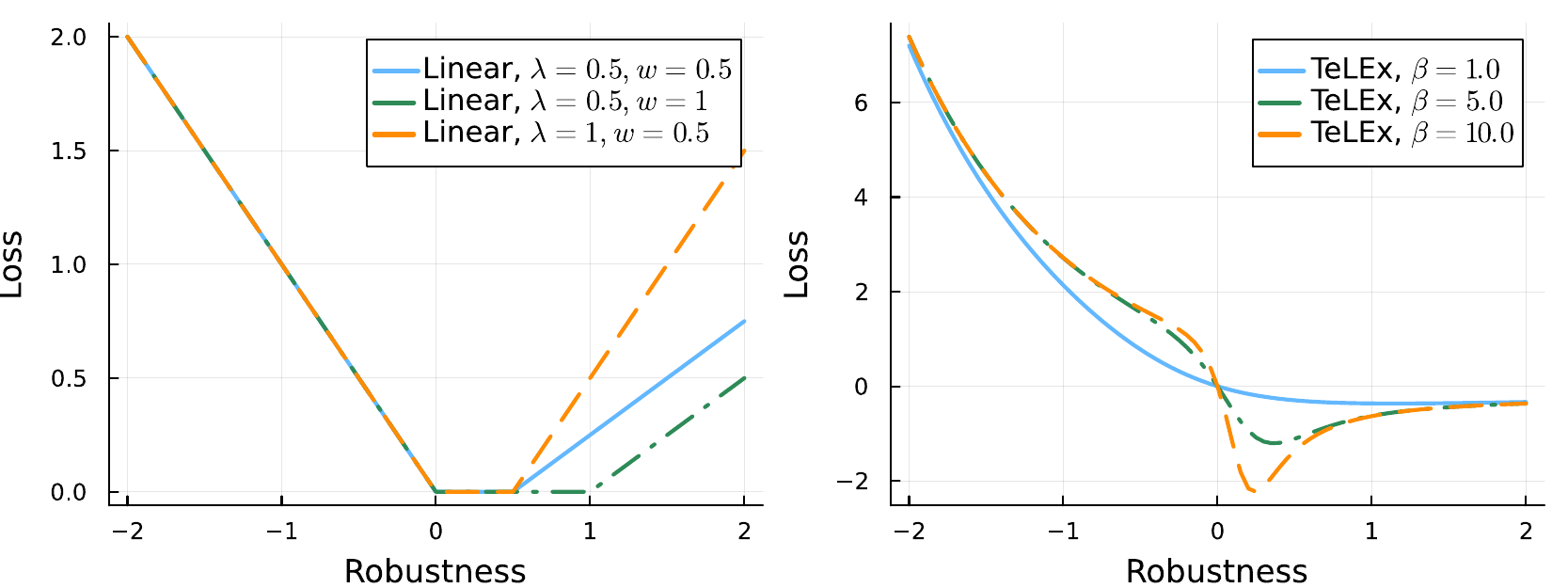}
	\caption{Comparison of the linear and TeLEx loss functions.}
	\label{fig:loss}
\vspace{-1em}
\end{wrapfigure}

\paragraph*{Loss function.} We tested two loss functions over STL robustness: a linear function \eqref{eq:linear-penalty} in which the loss is 0 in some interval $[0, w]$ (we used $w=0.5$) and the continuous loss \eqref{eq:telex-penalty} developed for the TeLEx learning algorithm (\cite{jha2019telex}). Both of these functions penalize negative and large positive robustness values. Negative values correspond to predicates that don't accurately describe the trajectory data, while large positive values are used as a proxy for overly-general (thus uninformative) predicates (\cite{jha2019telex}). However, the two penalties differ in their range of values and design (Figure \ref{fig:loss}).
To adapt a penalty over a single robustness value for an interval $[l,h]\subset\R$, we simply sum the penalties. To ensure $l<h$ for the TeLEx penalty, we subtract an offset $w$ from $h$ (we use the same value: $w=0.5$).
\begin{align}
\ell_{\text{linear}}(l,h) =& \max\{\max\{-l, 0, \lambda(l-w)\}, \max\{-h, 0, \lambda(h - w)\}\}
\label{eq:linear-penalty}
\\
\ell_{\text{TeLEx}}(l,h) =& -\frac{1}{l + e^{\left(-\beta_l l\right)}} + e^{-l} - \frac{1}{(h-w) + e^{\left(-\beta_h (h-w)\right)}} + e^{-(h-w)}
\label{eq:telex-penalty}
\end{align}

We also introduce two penalties computed directly on the expression tree representing predicate $\phi$.
The length penalty $p_{\text{length}}(\phi)$ 
	is defined as the number of nodes in the tree. This penalty is a standard tactic in genetic programming.
The trivial penalty $p_{\text{trivial}}(\phi) = \texttt{len}(\phi) - \texttt{len}(\texttt{cnf}(\phi))$ penalizes complex predicates that can be simplified. This prevents our algorithm from learning predicates that simplify to the trivial expressions $\top$ and $\bot$. The function \verb|cnf| simplifies $\phi$ to conjunctive normal form and is implemented using the Z3 solver \cite{z3}.
We then use grammar $\Gg_m$ and an expression optimization algorithm to learn a predicate $\phi$ that minimizes the loss \eqref{eq:loss-final} where $a_1$ and $a_2$ control the relative weight of each penalty.
\begin{equation}
L(\phi) = \ell(l_\phi,h_\phi) + a_1 p_{\text{length}}(\phi) + a_2 p_{\text{trivial}}(\phi)
\label{eq:loss-final}
\end{equation}
\section{Results}
\begin{wrapfigure}{r}{0.3\textwidth}
\centering
\vspace{-1em}
	\includegraphics[width=1.0\linewidth]{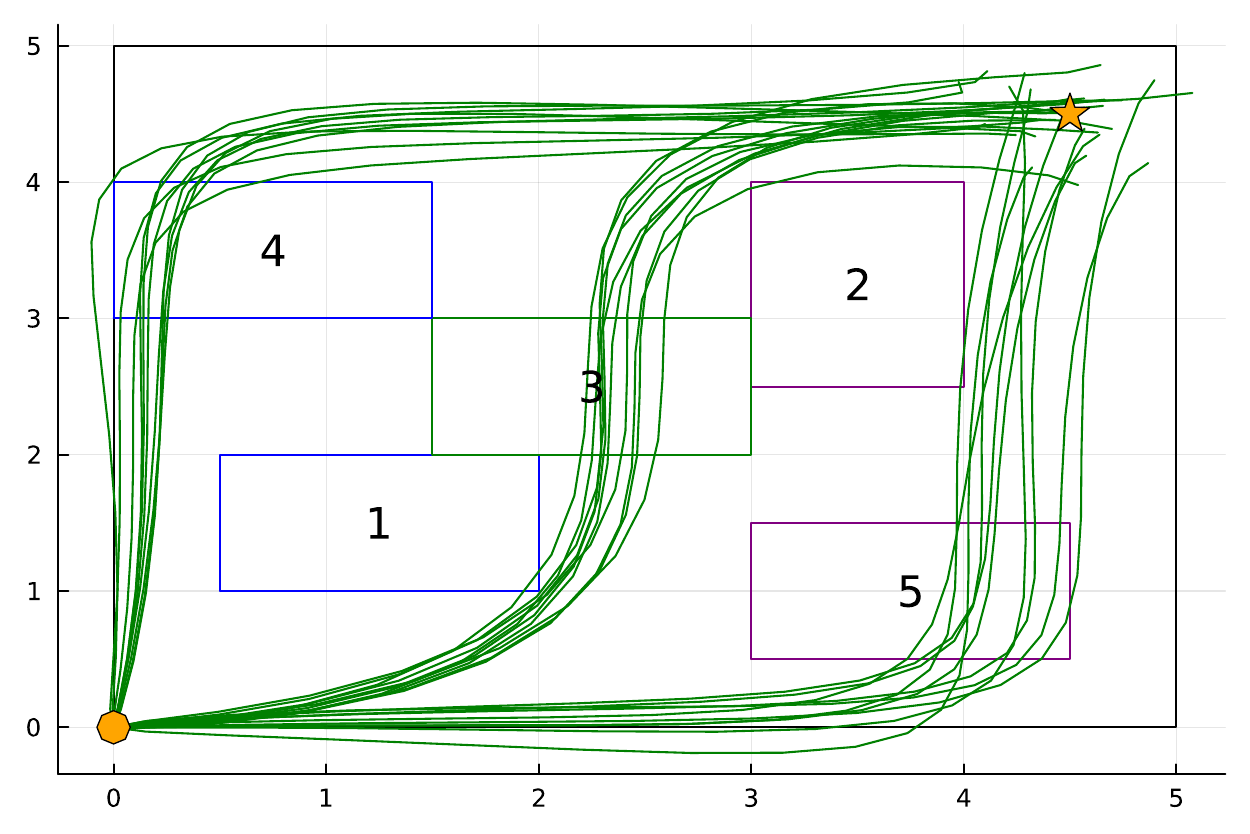}
	\caption{Trajectory data.} 
	\label{fig:smalldata}
\end{wrapfigure}
We demonstrate our method on a synthetic dataset $\mathcal{D}$ containing $N=2000$ ground-truth 2D position trajectories with 20 waypoints each (Figure \ref{fig:smalldata}). For each trajectory, we use a noisy observation of the waypoints to generate 100 predictions. In all experiments we used an even five-way split, reshuffling the data between trials.
We define $m = 5$ atoms where $\phi_i$ denotes the temporal predicate $\phi_i = \Eventually(\texttt{inside\_box}_i)$. The Boolean predicate $\texttt{inside\_box}_i$ constrains the 2-D state $(x_{t,1}, x_{t,2})$ to a box defined by two points $(a_1,b_1)$ and $(a_2,b_2)$:
\begin{equation*}
\texttt{inside\_box}_i(x_t) = (x_{t,1} \geq a_1 \wedge x_{t,1}\leq a_2) \wedge (x_{t,2} \geq b_1 \wedge x_{t,2}\leq b_2).
\end{equation*}
For each atom $\phi_i$, we train $f^1_{\phi_i}$ and $f^2_{\phi_i}$ using k-nearest neighbors (kNN) regression to predict the $\nicefrac{\alpha}{2}$ and $1-\nicefrac{\alpha}{2}$ quantiles of $\rho_{\phi_i}$ (\cite{knnconformal2011}). Using $f^1_{\phi_i}, f^2_{\phi_i},\ i=1,\dots,M$, we precompute conformalized CIs on $\mathcal{D}_{\test}$. We then use these intervals in our expression optimization step to search for a predicate $\phi^\star$ in grammar $\mathcal{G}$ that minimizes the loss \eqref{eq:loss-final}.

\begin{figure}[ht]
	\centering
    \begin{minipage}{0.65\textwidth}
	\includegraphics[width=\linewidth]{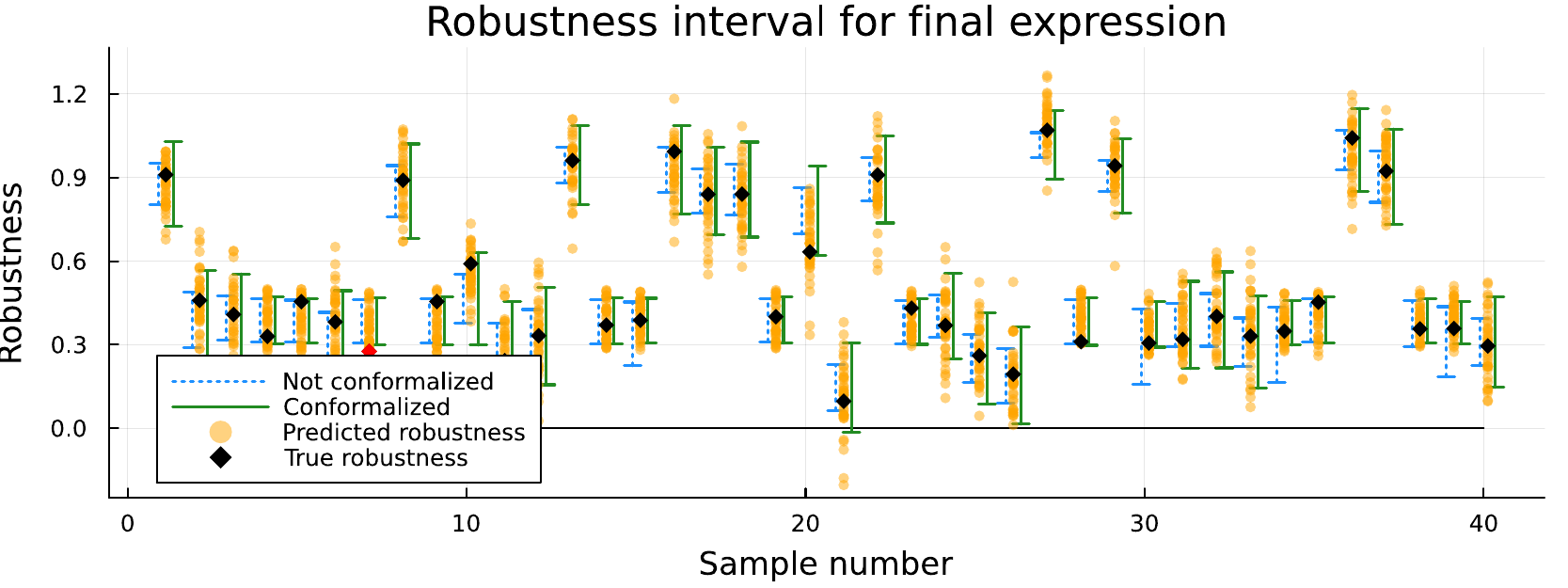}
    \end{minipage}
        \begin{minipage}{0.34\textwidth}
\begin{verbnobox}[\fontsize{7pt}{7pt}\selectfont]
and
| or
| | Eventually(inside_box(2))
| | Eventually(inside_box(2))
| | not
| | | Eventually(inside_box(1))
| or
| | not
| | | Eventually(inside_box(1))
| | Eventually(inside_box(4))
| | Eventually(inside_box(3))
\end{verbnobox}
\normalsize
\end{minipage}

	\caption{Example predicate $\phi^\star$ and confidence interval predictions for 40 randomly sampled trajectories from $\Dd_{\val}$. We plot a scatterplot of predicted robustness values for each sample; the predicted intervals generated by $f^1_{\phi^\star}$ and $f^2_{\phi^\star}$; and the true robustness for each sample. The true robustness is shown in red if it lies outside the predicted conformal CI.}
	\label{fig:robustness}
	\vspace{-1em}
\end{figure}

As there are multiple valid predicates to describe our sample dataset, we provide several examples of $\phi^\star$ collected over different trials. To empirically validate the performance of our algorithm and establish statistical significance, we performed 50 trials for each algorithm configuration, reshuffling the train, test, calibration, and validation splits and retraining the predictors for each trial. We report results with $2\sigma$ error bars for different configurations of our algorithm and several ablation studies (Tables \ref{tab:optimizers} and \ref{tab:ablation}). Predicted confidence intervals for $\rho_{\phi^\star}$ from a single trial are presented in Figure \ref{fig:robustness}.

\paragraph*{Metrics}
We used several metrics to evaluate our algorithm. The \textit{efficiency} is the average CI width: a standard metric (\cite{romano2019conformalized}).
The \textit{error rate} is the fraction of true robustness values $\rho_{\phi^\star}(x^{(i)})$ that lie outside their predicted CIs. (\cite{cairoli2023conformal}).
The \textit{trival rate} is the percentage of learned predicates that simplify to $\top$ or $\bot$.
The \textit{negative percentage} is defined as $\text{length}([l,h] \cap (-\infty, 0))/\text{length}([l,h])$; the percentage of interval $[l,h]$ that overlaps with the negative reals. A nonzero percentage corresponds to a CI containing negative robustness values, indicating the learned predicate doesn't accurately describe the dataset.
Additionally, we report the average values of $l_{\phi^\star}$ and $h_{\phi^\star}$ and the error rate before conformalizing the quantile predictors for $\phi^\star$\footnote{We report this primarily to illustrate the value of the CQR procedure in enforcing the desired error rate.}.

\subsection{Expression optimization algorithms}
We tested several different algorithms implemented by ExprOptimization.jl (\cite{exproptimization}) to show that our approach is robust to the choice of optimizer. Genetic programming and grammatical evolution are both randomized evolution-based algorithms for learning tree-based structures. The Monte Carlo method samples random expression trees and keeps the best-performing one, and the cross-entropy method optimizes a probability distribution of expression trees with the lowest losses. All of these algorithms require only a grammar $\Gg$ and a loss function over expression trees.
The genetic programming and grammatical evolution algorithms show a statistically significant improvement in the percentage of negative intervals over Monte Carlo and cross-entropy (Table \ref{tab:optimizers}). 

\begin{table*}
	\caption{Comparison of different expression optimization methods. We conducted 50 trials over $\Dd$ containing 1,600 trajectories. We used the parameters $\beta=5, a_1=0.001, a_2=1.0$. For the Monte Carlo method, we used 10,000 samples. For the other algorithms, which are iterative, we used 500 iterations. In all tests, $\alpha = 0.1$.}
	\label{tab:optimizers}
 \centering
  \resizebox{\textwidth}{!}{%
	\begin{tabular}{@{} p{2.5cm}p{2.6cm}p{2.4cm}p{2.4cm}p{1cm}p{2.4cm}p{2.4cm}p{2.0cm}p{2.0cm}@{}}
		\toprule
Trial &		Error rate (nonconformal) & Error rate (conformal) & Efficiency (conformal) & Trivial rate & Mean $l_{\phi^\star}$ & Mean $h_{\phi^\star}$ & Negative

percentage & Exec time (s)
\\
\midrule

Genetic Prog. & 0.155 $\pm$ 0.110 & 0.099 $\pm$ 0.041 & 0.257 $\pm$ 0.072 & 0.1\% & \textbf{0.471 $\pm$ 0.482} & \textbf{0.728 $\pm$ 0.496} & \textbf{0.1 $\pm$ 0.6} & 49.9 $\pm$ 2.1
\\
Monte Carlo & 0.153 $\pm$ 0.110 & 0.106 $\pm$ 0.045 & 0.253 $\pm$ 0.097 & 0.0 & 0.265 $\pm$ 0.311 & 0.517 $\pm$ 0.273 & 4.7 $\pm$ 18.0 & 76.3 $\pm$ 18.9
\\
Gramm. Evol. & 0.159 $\pm$ 0.080 & 0.103 $\pm$ 0.050 & \textbf{0.264 $\pm$ 0.069} & 0.0 & 0.426 $\pm$ 0.606 & 0.690 $\pm$ 0.602 & 0.8 $\pm$ 3.2 & 50.0 $\pm$ 9.8
\\
Cross Entropy & 0.147 $\pm$ 0.117 & 0.100 $\pm$ 0.051 & 0.210 $\pm$ 0.108 & 0.0 & 0.192 $\pm$ 0.392 & 0.447 $\pm$ 0.375 & 13.7 $\pm$ 32.0 & 66.1 $\pm$ 0.95
\\
\bottomrule
	\end{tabular}
 }
\vspace{-1em}
\end{table*}

\subsection{Ablation testing and comparison to baseline}
We conducted several ablation tests, presented in Table \ref{tab:ablation}. First, we see that the TeLEx loss function with parameter $\beta=5$ provides a statistically significant improvement in the negative interval percentage compared to a simple linear loss. The trivial penalty provides the most dramatic improvement, reducing the rate of trivial expressions from 68\% to 0\%. 
Finally, the ``No intervals'' test is a baseline expression optimization approach that attempts to learn an STL predicate from the predicted robustness values, then trains a CQR predictor on the resulting $\phi^\star$. Although this test has a similar negative interval percentage to our approach, the average $l_{\phi^\star}$ and $h_{\phi^\star}$ are lower. We conclude that carrying the confidence interval information through the expression optimization step enables our algorithm to find more robust predicates, possibly because the interval data is less vulnerable to outliers.
\begin{table*}[t]
	\caption{Results of ablation testing. We conducted 50 trials over $\mathcal{D}$ containing 1,600 trajectories. We used the parameters $a_1=0.001, a_2=1.0, \beta=5$ and the genetic programming algorithm with 500 iterations. In all tests, $\alpha = 0.1$.}
	\label{tab:ablation}
 \centering
  \resizebox{\textwidth}{!}{%
	\begin{tabular}{@{} p{2.5cm}p{2.6cm}p{2.4cm}p{2.4cm}p{1cm}p{2.4cm}p{2.4cm}p{2.0cm}p{2.0cm}@{}}
		\toprule
Trial &		Error rate (nonconformal) & Error rate (conformal) & Efficiency (conformal) & Trivial rate & Mean $l_{\phi^\star}$ & Mean $h_{\phi^\star}$ & Negative

percentage & Exec time (s)
\\
\midrule
TeLEx loss & 0.155 $\pm$ 0.110 & 0.099 $\pm$ 0.041 & 0.257 $\pm$ 0.072 & 0.1\% & \textbf{0.471 $\pm$ 0.482} & \textbf{0.728 $\pm$ 0.496} & \textbf{0.1 $\pm$ 0.6} &  49.9 $\pm$ 2.1
\\
Linear loss & 0.200 $\pm$ 0.058 & 0.101 $\pm$ 0.046 & 0.259 $\pm$ 0.031  & \textbf{0\%} & 0.160 $\pm$ 0.071 &  0.419 $\pm$ 0.047 & 4.0 $\pm$ 3.3 & 92.0 $\pm$ 68.5
\\
No $p_{\text{trivial}}$ & 0.080 $\pm$ 0.054 & 0.104 $\pm$ 0.043 & 0.274 $\pm$ 0.037 & 60\% & \textbf{0.588 $\pm$  0.207} & \textbf{0.862 $\pm$  0.181} & \textbf{0.0 $\pm$ 0.0} & 3.5 $\pm$ 1.1
\\
No intervals & N/A & 0.101 $\pm$ 0.045 & 0.250 $\pm$ 0.054 & 0\% & 0.249 $\pm$  0.121 & 0.499 $\pm$ 0.118 & 0.5 $\pm$ 1.8 & 85.5 $\pm$ 12.7\\
\bottomrule
	\end{tabular}
 }
\vspace{-1em}
\end{table*}

\vspace{-0.5em}
\subsection{VRU dataset}
\begin{wrapfigure}{R}{0.4\textwidth}
    \centering
    \includegraphics[width=\linewidth]{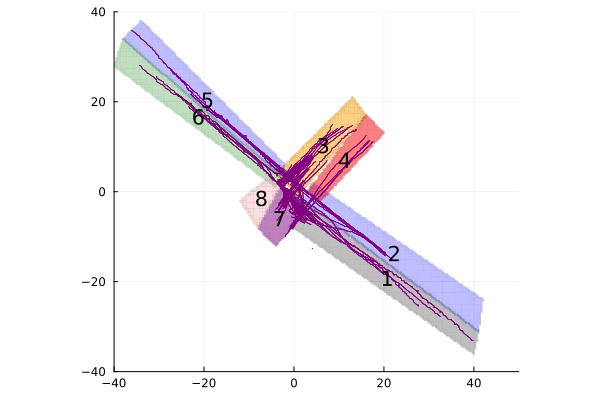}
    \caption{Samples from the VRU dataset overlaid on 8 regions used to define STL atoms.}
    \label{fig:map}
\end{wrapfigure}
\begin{figure}[ht]
	\centering
\begin{minipage}{0.65\textwidth}
    \includegraphics[width=1.0\linewidth]{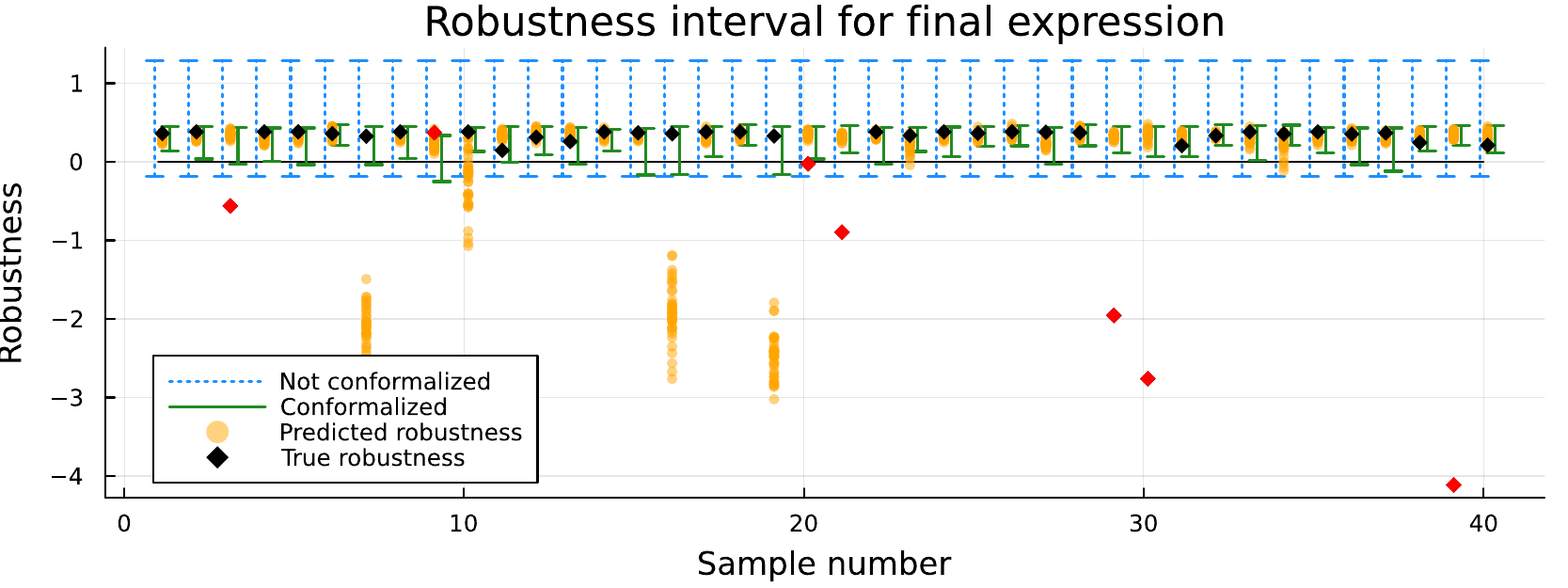}
\end{minipage}
    \begin{minipage}{0.34\textwidth}
\begin{verbnobox}[\fontsize{6pt}{6pt}\selectfont]
and
| or
| | not
| | | And(Always(is_bike, Always(in_6))
| | And(Always(is_ped), Always(in_3))
| or
| | And(Always(is_ped), Always(in_3))
| | And(Always(is_ped), Always(in_7))
| | And(Always(is_ped), Always(in_8))
| | And(Always(is_bike), Always(in_8))
\end{verbnobox}
\normalsize
\end{minipage}

	\caption{Example predicate $\phi^\star$ and confidence interval predictions for 40 randomly sampled trajectories from the VRU dataset using the same visualization as in Figure \ref{fig:robustness}. The red points are true robustness values outside their CQR intervals.}
	\label{fig:robustness-vru}
	\vspace{-1em}
\end{figure}

We also tested our algorithm on the VRU dataset (\cite{VRUDataset}) containing 1068 pedestrian and 464 cyclist trajectories recorded at an urban intersection. Each trajectory includes the time, $x$ and $y$ position, whether the agent is a pedestrian or bicycle, and whether the agent is starting, stopping, moving, or waiting; thus our state vector has 4 components. We identified eight regions of interest, shown in Figure \ref{fig:map}. For ease of development, the regions are represented by linear constraints; however there is no technical limitation preventing the use of other shapes. We defined a set of 16 atoms describing both the type of agent (pedestrian or bicycle) and what region they are \textit{Always} inside, using $\alpha=0.1$ and the genetic programming algorithm for expression optimization. We did not modify any hyperparameters.
Our algorithm successfully finds predicates that describe the data with the desired probability (Figure \ref{fig:robustness-vru}). These results show that our work is applicable to real datasets and further, that it can perform well without parameter tuning on each dataset of interest.

\section{Limitations}
One limitation of our algorithm is the restriction to the simpler grammar \eqref{eq:grammar} consisting of predefined temporal predicates and Boolean operators. Future work may expand the grammar to allow for combining predicates using temporal operators or optimizing the interval length of temporal predicates (the range of time steps the predicate applies to).
Another limitation is the use of atoms, which serve as features of the data expressed in the STL language. In practice, our choice of atoms will be informed by our knowledge of the system being studied. There are both benefits and drawbacks to this approach. In many cases system designers will have clear ideas about behaviors that may be present in the dataset -- for example, in a driving scenario one expects to see lane changes and left or right turns and could construct atoms describing these behaviors, ensuring the learned predicate will be presented in a useful format. However, selecting atoms also risks leaving out features we don't expect to observe: for example, illegal turns or unsafe maneuvers in a driving scenario.
\vspace{-0.5em}
\section{Conclusion}
We have developed a novel algorithm to fit STL predicates to data with statistical guarantees. We show that genetic programming or grammatical evolution, both randomized methods for learning expression trees, perform well in this task and that our approach, which optimizes explicitly over confidence intervals of robustness values, compares favorably to post-hoc calibration of a predicate learned using a naive approach.
Because our method provides a statistical guarantee on the STL robustness under well-understood conditions (exchangeability of trajectory data), future work in safety-critical applications can leverage our algorithm for formal verification of systems from data. Moreover, because our method takes as input an \textit{observation} of a trajectory, it has applications in robotic autonomy where, for example, an agent may need to complete a motion planning task with limited observations of other agents' trajectories. An STL predicate that describes agents' predicted future trajectories could be used to constrain this motion planner and avoid unsafe situations. Future work could include improving the speed of the algorithm for real-time applications, as well as expanding the capabilities of our expression optimization approach.
\vspace{-0.5em}
\acks{
Professor Mykel Kochenderfer provided the insight to compare different expression optimization algorithms, as well as valuable comments on the presentation of our algorithm and results. Nick Landolfi and Aaron Mishkin provided feedback on drafts of this paper. Professor Lars Lindemann identified an error in a preprint of this paper, which has since been corrected. This research was supported, in part, by Ford Motor Co. under the Stanford-Ford Alliance, agreement number 235158.}

\bibliography{bibliography}

\begin{thebibliography}{30}
\providecommand{\natexlab}[1]{#1}
\providecommand{\url}[1]{\texttt{#1}}
\expandafter\ifx\csname urlstyle\endcsname\relax
  \providecommand{\doi}[1]{doi: #1}\else
  \providecommand{\doi}{doi: \begingroup \urlstyle{rm}\Url}\fi

\bibitem[VRU(2020)]{VRUDataset}
{VRU} {T}rajectory {D}ataset, 2020.
\newblock \url{https://www.th-ab.de/hochschule/organisation/organisationseinheiten/labor-fuer-kooperative-automatisierte-verkehrssysteme/trajectory-dataset/}.

\bibitem[Angelopoulos et~al.(2023)Angelopoulos, Bates, Fisch, Lei, and Schuster]{angelopoulos2023conformal}
Anastasios~N. Angelopoulos, Stephen Bates, Adam Fisch, Lihua Lei, and Tal Schuster.
\newblock Conformal risk control, 2023.

\bibitem[Bombara and Belta(2021)]{bombara2021decision}
Giuseppe Bombara and Calin Belta.
\newblock Offline and online learning of signal temporal logic formulae using decision trees.
\newblock \emph{ACM Trans. Cyber-Phys. Syst.}, 5\penalty0 (3), 3 2021.
\newblock ISSN 2378-962X.
\newblock \doi{10.1145/3433994}.
\newblock URL \url{https://doi.org/10.1145/3433994}.

\bibitem[Buzhinsky(2019)]{8972130}
Igor Buzhinsky.
\newblock Formalization of natural language requirements into temporal logics: a survey.
\newblock In \emph{2019 IEEE 17th International Conference on Industrial Informatics (INDIN)}, volume~1, pages 400--406, 2019.
\newblock \doi{10.1109/INDIN41052.2019.8972130}.

\bibitem[Cairoli et~al.(2023)Cairoli, Paoletti, and Bortolussi]{cairoli2023conformal}
Francesca Cairoli, Nicola Paoletti, and Luca Bortolussi.
\newblock Conformal quantitative predictive monitoring of {STL} requirements for stochastic processes.
\newblock In \emph{Proceedings of the 26th ACM International Conference on Hybrid Systems: Computation and Control}, pages 1--11, 2023.

\bibitem[Corso and Kochenderfer(2020)]{corso2020interpretable}
Anthony Corso and Mykel~J. Kochenderfer.
\newblock Interpretable safety validation for autonomous vehicles, 2020.

\bibitem[De~Moura and Bj\o{}rner(2008)]{z3}
Leonardo De~Moura and Nikolaj Bj\o{}rner.
\newblock Z3: an efficient smt solver.
\newblock In \emph{Proceedings of the Theory and Practice of Software, 14th International Conference on Tools and Algorithms for the Construction and Analysis of Systems}, TACAS'08/ETAPS'08, page 337–340, Berlin, Heidelberg, 2008. Springer-Verlag.
\newblock ISBN 3540787992.

\bibitem[Donz{\'e} and Maler(2010)]{donze2010robust}
Alexandre Donz{\'e} and Oded Maler.
\newblock Robust satisfaction of temporal logic over real-valued signals.
\newblock In \emph{International Conference on Formal Modeling and Analysis of Timed Systems}, pages 92--106. Springer, 2010.

\bibitem[Hahn et~al.(2022)Hahn, Schmitt, Tillman, Metzger, Siber, and Finkbeiner]{hahn2022formal}
Christopher Hahn, Frederik Schmitt, Julia~J. Tillman, Niklas Metzger, Julian Siber, and Bernd Finkbeiner.
\newblock Formal specifications from natural language, 2022.

\bibitem[Jha et~al.(2019)Jha, Tiwari, Seshia, Sahai, and Shankar]{jha2019telex}
Susmit Jha, Ashish Tiwari, Sanjit~A Seshia, Tuhin Sahai, and Natarajan Shankar.
\newblock Telex: learning signal temporal logic from positive examples using tightness metric.
\newblock \emph{Formal Methods in System Design}, 54:\penalty0 364--387, 2019.

\bibitem[Koymans(1990)]{koymans1990specifying}
Ron Koymans.
\newblock Specifying real-time properties with metric temporal logic.
\newblock \emph{Real-time systems}, 2\penalty0 (4):\penalty0 255--299, 1990.

\bibitem[Koza(1994)]{koza1994genetic}
John~R Koza.
\newblock Genetic programming as a means for programming computers by natural selection.
\newblock \emph{Statistics and computing}, 4:\penalty0 87--112, 1994.

\bibitem[Lee and Kochenderfer(2021)]{exproptimization}
Ritchie Lee and Mykel Kochenderfer.
\newblock Expr{O}ptimization.jl.
\newblock https://github.com/sisl/ExprOptimization.jl, 2021.

\bibitem[Lei et~al.(2018)Lei, G’Sell, Rinaldo, Tibshirani, and Wasserman]{lei2018distribution}
Jing Lei, Max G’Sell, Alessandro Rinaldo, Ryan~J Tibshirani, and Larry Wasserman.
\newblock Distribution-free predictive inference for regression.
\newblock \emph{Journal of the American Statistical Association}, 113\penalty0 (523):\penalty0 1094--1111, 2018.

\bibitem[Leung et~al.(2023)Leung, Ar{\'e}chiga, and Pavone]{leung2023backpropagation}
Karen Leung, Nikos Ar{\'e}chiga, and Marco Pavone.
\newblock Backpropagation through signal temporal logic specifications: Infusing logical structure into gradient-based methods.
\newblock \emph{The International Journal of Robotics Research}, 42\penalty0 (6):\penalty0 356--370, 2023.

\bibitem[Lindemann et~al.(2023)Lindemann, Qin, Deshmukh, and Pappas]{lindemann2023conformal}
Lars Lindemann, Xin Qin, Jyotirmoy~V Deshmukh, and George~J Pappas.
\newblock Conformal prediction for {STL} runtime verification.
\newblock In \emph{Proceedings of the ACM/IEEE 14th International Conference on Cyber-Physical Systems (with CPS-IoT Week 2023)}, pages 142--153, 2023.

\bibitem[Maierhofer et~al.(2022)Maierhofer, Moosbrugger, and Althoff]{maierhofer2022formalization}
Sebastian Maierhofer, Paul Moosbrugger, and Matthias Althoff.
\newblock Formalization of intersection traffic rules in temporal logic.
\newblock In \emph{2022 IEEE Intelligent Vehicles Symposium (IV)}, pages 1135--1144. IEEE, 2022.

\bibitem[Maler and Nickovic(2004)]{maler2004monitoring}
Oded Maler and Dejan Nickovic.
\newblock Monitoring temporal properties of continuous signals.
\newblock In \emph{International Symposium on Formal Techniques in Real-Time and Fault-Tolerant Systems}, pages 152--166. Springer, 2004.

\bibitem[Nenzi(2023)]{nenzi2023learning}
Laura Nenzi.
\newblock Learning temporal logic formulas from time-series data (invited talk).
\newblock In \emph{30th International Symposium on Temporal Representation and Reasoning (TIME 2023)}. Schloss Dagstuhl-Leibniz-Zentrum f{\"u}r Informatik, 2023.

\bibitem[Nenzi et~al.(2018)Nenzi, Silvetti, Bartocci, and Bortolussi]{nenzi2018robust}
Laura Nenzi, Simone Silvetti, Ezio Bartocci, and Luca Bortolussi.
\newblock A robust genetic algorithm for learning temporal specifications from data, 2018.

\bibitem[Papadopoulos et~al.(2011)Papadopoulos, Vovk, and Gammerman]{knnconformal2011}
H.~Papadopoulos, V.~Vovk, and A.~Gammerman.
\newblock Regression conformal prediction with nearest neighbours.
\newblock \emph{Journal of Artificial Intelligence Research}, 40:\penalty0 815–840, April 2011.
\newblock ISSN 1076-9757.
\newblock \doi{10.1613/jair.3198}.
\newblock URL \url{http://dx.doi.org/10.1613/jair.3198}.

\bibitem[Pigozzi et~al.(2021)Pigozzi, Medvet, and Nenzi]{pigozzi2021mining}
Federico Pigozzi, Eric Medvet, and Laura Nenzi.
\newblock Mining road traffic rules with signal temporal logic and grammar-based genetic programming.
\newblock \emph{Applied Sciences}, 11\penalty0 (22):\penalty0 10573, 2021.

\bibitem[Qin and Deshmukh(2020)]{qin2020clairvoyant}
Xin Qin and Jyotirmoy~V Deshmukh.
\newblock Clairvoyant monitoring for signal temporal logic.
\newblock In \emph{Formal Modeling and Analysis of Timed Systems: 18th International Conference, FORMATS 2020, Vienna, Austria, September 1--3, 2020, Proceedings 18}, pages 178--195. Springer, 2020.

\bibitem[Qin et~al.(2022)Qin, Xia, Zutshi, Fan, and Deshmukh]{qin2022statistical}
Xin Qin, Yuan Xia, Aditya Zutshi, Chuchu Fan, and Jyotirmoy~V Deshmukh.
\newblock Statistical verification of cyber-physical systems using surrogate models and conformal inference.
\newblock In \emph{2022 ACM/IEEE 13th International Conference on Cyber-Physical Systems (ICCPS)}, pages 116--126. IEEE, 2022.

\bibitem[Raman et~al.(2015)Raman, Donz\'{e}, Sadigh, Murray, and Seshia]{reactive_synthesis}
Vasumathi Raman, Alexandre Donz\'{e}, Dorsa Sadigh, Richard~M. Murray, and Sanjit~A. Seshia.
\newblock Reactive synthesis from signal temporal logic specifications.
\newblock In \emph{Proceedings of the 18th International Conference on Hybrid Systems: Computation and Control}, HSCC '15, page 239–248, New York, NY, USA, 2015. Association for Computing Machinery.
\newblock ISBN 9781450334334.
\newblock \doi{10.1145/2728606.2728628}.
\newblock URL \url{https://doi.org/10.1145/2728606.2728628}.

\bibitem[Romano et~al.(2019)Romano, Patterson, and Candes]{romano2019conformalized}
Yaniv Romano, Evan Patterson, and Emmanuel Candes.
\newblock Conformalized quantile regression.
\newblock \emph{Advances in neural information processing systems}, 32, 2019.

\bibitem[Roy et~al.(2023)Roy, Gaglione, Baharisangari, Neider, Xu, and Topcu]{roy2023learning}
Rajarshi Roy, Jean-Rapha{\"e}l Gaglione, Nasim Baharisangari, Daniel Neider, Zhe Xu, and Ufuk Topcu.
\newblock Learning interpretable temporal properties from positive examples only.
\newblock In \emph{Proceedings of the AAAI Conference on Artificial Intelligence}, volume~37, pages 6507--6515, 2023.

\bibitem[Vazquez-Chanlatte et~al.(2017)Vazquez-Chanlatte, Deshmukh, Jin, and Seshia]{vazquez2017logical}
Marcell Vazquez-Chanlatte, Jyotirmoy~V Deshmukh, Xiaoqing Jin, and Sanjit~A Seshia.
\newblock Logical clustering and learning for time-series data.
\newblock In \emph{Computer Aided Verification: 29th International Conference, CAV 2017, Heidelberg, Germany, July 24-28, 2017, Proceedings, Part I 30}, pages 305--325. Springer, 2017.

\bibitem[Vovk et~al.(2005)Vovk, Gammerman, and Shafer]{vovk2005algorithmic}
Vladimir Vovk, Alexander Gammerman, and Glenn Shafer.
\newblock \emph{Algorithmic learning in a random world}, volume~29.
\newblock Springer, 2005.

\bibitem[Zhao et~al.(2024)Zhao, Hoxha, Fainekos, Deshmukh, and Lindemann]{zhao2024robustconformalpredictionstl}
Yiqi Zhao, Bardh Hoxha, Georgios Fainekos, Jyotirmoy~V. Deshmukh, and Lars Lindemann.
\newblock Robust conformal prediction for stl runtime verification under distribution shift, 2024.
\newblock URL \url{https://arxiv.org/abs/2311.09482}.

\end{thebibliography}

\end{document}